# Extracting an English-Persian Parallel Corpus from Comparable Corpora

Akbar Karimi[*], Ebrahim Ansari[*], Bahram Sadeghi Bigham[*]
[*] Department of Computer Sciences and Information Technology, Institute for Advanced Studies in Basic Sciences
(IASBS), Zanjan, Iran
{ak.karimi, ansari, b_sadeghi_b}@iasbs.ac.ir

**Abstract**
Parallel data are an important part of a reliable Statistical Machine Translation (SMT) system. The more of these data are available, the better the quality of the SMT system. However, for some language pairs such as Persian-English, parallel sources of this kind are scarce. In this paper, a bidirectional method is proposed to extract parallel sentences from English and Persian document aligned Wikipedia. Two machine translation systems are employed to translate from Persian to English and the reverse after which an IR system is used to measure the similarity of the translated sentences. Adding the extracted sentences to the training data of the existing SMT systems is shown to improve the quality of the translation. Furthermore, the proposed method slightly outperforms the one-directional approach. The extracted corpus consists of about 200,000 sentences which have been sorted by their degree of similarity calculated by the IR system and is freely available for public access on the Web[1].

**Keywords:** Parallel Sentence Extraction, Comparable Corpora, Statistical Machine Translation, Wikipedia, English-Persian Corpus

## 1. Introduction

Due to the abundance of data on the Internet, statistical machine translation (SMT) has gained more popularity. In order to build an SMT system, parallel corpora are of high importance. These parallel resources which have been aligned on the sentence level in two languages (source and target), are used in the training phase of the SMT system. Therefore, the larger the parallel corpora are, the better the performance of the SMT system is. However, for some language pairs such as Persian-English not much data of this type is available. This lack of parallel data has led researchers to make use of other available data called comparable corpora which contain a mixture of parallel and partially parallel sentences. They can be given a certain degree of comparability which ranges from lowly comparable to highly comparable (Li and Gaussier, 2010). Research shows that using these corpora can help improve the performance of the SMT system.

There are several sources such as news articles, company manuals, Wikipedia articles, and so forth which can be considered as comparable corpora. In this work, our aim is to extract parallel sentences for Persian-English language pair from Wikipedia documents using a new approach to improve the Persian-English SMT system.

Our method consists of two main parts: translation and information retrieval. For the translation part (Persian to English and English to Persian), we employed Moses Toolkit developed by Koehn et al., (2007) which is an open-source toolkit developed for phrase-based translation and for the IR step we utilized the Lucene IR system[2]. Lucene has been designed to work with queries which are fed into the system one by one and the results shown by the IR system for a query are numbers representing the degree of the documents' relevance to the query. To compute the similarity of two sentences, Lucene's original source code was modified so that the queries could be read from a text file and the most relevant sentences from another file could be given as the result of each query by the IR system. To carry out our experiments, we needed documents in Persian and English whose topics were the same. Therefore, we downloaded the document aligned Persian-English Wikipedia from Linguatools[3]. It is an XML file that contains the English documents for each of which there is a Persian entry. There are 363183 document pairs in this file.

The rest of the paper is as follows: In Section 2, a review of some of the related work is presented. Section 3 is dedicated to describing our method. Then, a detailed explanation of our experiments and their results is given in Section 4 and the final section concludes the paper.

## 2. Related Work

There have been several papers written on the use of comparable corpora. Due to the lack of enough parallel data for many language pairs, some have also proposed using English as a pivot language to extract parallel resources and for translation purposes.

Resnik et al., (2003), working on web pages, use STRAND, which is their structural filtering system, to recognize parallel pairs. In order to do so, they specify a set of pair-specific values and experiment on English-Chinese corpus, reporting precision and recall of 98 percent and 61 percent, respectively.

Koehn et al., (2005) extract parallel texts for 11 languages from the proceedings of the European Parliament to be used as the training data for building SMT systems. Smith et al., (2010) work on the document level aligned Wikipedia data for three language pairs, Spanish-English, Bulgarian-English, and German-English, and using Hidden Markov Model for word alignment, they extract parallel sentences for the aforementioned language pairs and build improved SMT systems.

Using a very small parallel corpus which contains only 100 thousand words and a bilingual dictionary, Munteanu and Marcu (2005) train a maximum entropy classifier to extract parallel sentences from large comparable corpora. They work with Arabic-English and French-English language pairs to carry out their experiments. In another work (Munteanu and Marcu, 2006), they extract sub-sentential

---

[1] https://iasbs.ac.ir/~ansari/nlp/pepc.html
[2] https://lucene.apache.org
[3] http://linguatools.org

fragments from non-parallel corpora that do not contain any parallelism on the sentence level.

Stefanescu and Ion (2013) work on Wikipedia to extract parallel sentences for English-German, English-Romanian, and English-Spanish language pairs. In order to find the parallel sentences from the comparable documents for each language pair, they make use of LEXACC, a tool developed by ACCURAT project for extraction of parallel sentences.

Do et al., (2010) propose a fully unsupervised method for parallel sentence extraction in which they build an SMT system using not parallel data but comparable data, and with this system, they translate the sentences from the source side of another comparable corpus to the target language. Then, they evaluate the translations by BLEU, NIST, and TER evaluation metrics, refeeding the ones that have been recognized as parallel into the SMT system and repeating the process. They claim that first few iterations of this process helps increase the number of parallel sentences resulting in improvements in the quality of the SMT system. In another work, using the aforementioned method and the English as the pivot language and a method called triangulation, Do et al., (2010) make an attempt to translate from Vietnamese to French.

Ansari et al., (2017) work on Persian-Italian languages using English as the pivot language. Sentences from Persian and Italian are translated into English and compared with each other by a new similarity metric which is based on Normalized Google Distance (NGD).

Linard et al., (2015) propose two approaches to bilingual lexicon extraction using English as the pivot language. One is to translate the source language to pivot and from that to the target language. The second approach is to translate both of the source and target languages into pivot language and then extract bilingual vocabulary.

Bakhshaei et al., (2015) introduce a generative model based on LDA concept to extract fragments and show that the baseline system with the additional fragments perform better than the baseline system alone.

Aker et al., (2013) use an SVM binary classifier for the extraction of bilingual terminology and they claim to have achieved an accuracy of 100% for the classifier. In another attempt to extract bilingual lexica, Seo et al., (2015) use self-organizing maps on comparable corpora for Korean-French and Korean-Spanish language pairs.

Using bootstrapping, Fung et al., (2004) work on very-non-parallel corpora and present a method for parallel sentence extraction, claiming that their method is 50% more effective than the baseline system. In their work, after matching the documents and extracting some parallel sentences, they rematch them based on the number of extracted parallel sentences and then carry out bootstrapping. The reason why they do this is due to a principle that they call "find-one-get-more" which means that if a sentence pair can be found in a document, more sentences are likely to exist in the same documents.

Rauf and Schwenk (2009, 2011) build an SMT system to translate one side of their bilingual corpus to be used as queries in an IR system to find their equivalents in the target language. To filter out the candidate sentences for each query, they use evaluation metrics such as word error rate (WER), translation error rate (TER), and translation error rate plus (TERp). They work with Arabic-English and French-English language pairs and report significant improvements in BLEU score.

## 3. Our Approach

When using a translation-based method to extract parallel sentences, the quality of the machine used for translation plays an important role. Since translating only one side of the corpus into another, which we call one-directional method, is not done flawlessly, it seems that if both sides were to be translated and used as queries, it would result in extracting better equivalents from the comparable corpus. We call this a bidirectional approach whose architecture is shown in Figure 1.

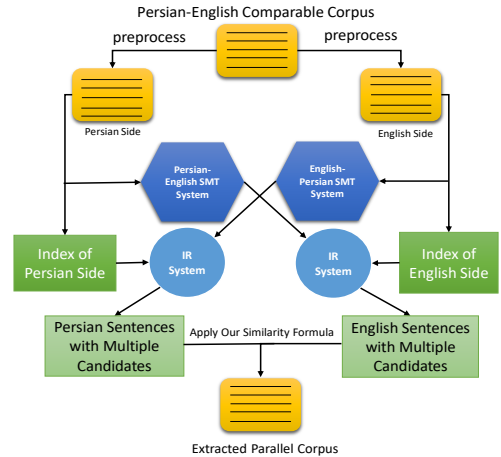

Figure 1: The architecture of the proposed bidirectional method for parallel sentence extraction.

As can be seen in the diagram, we first obtain the data from the Internet after which the data is translated in both directions and then parallel sentences are extracted from the candidate sentences found by Lucene IR system. Each of these steps is discussed in depth in the following sections.

### 3.1 Data Preparation

Wikipedia articles from which we want to extract parallel sentence have been made available by Linguatools website. The files are in XML file format containing all the documents in Wikipedia for many language pairs. The Persian-English corpus was downloaded for extracting parallel sentences. Since it is an XML file, it contains markup language, links, tables, figures, and so forth. Therefore, all the unnecessary characters in the file need to be removed first. This was carried out by writing a python script, and as a result, we obtained two plain texts containing only Persian and English sentences. In the process of obtaining plain texts from them, we ignored some documents and some sentences. If the number of sentences in a document was lower than 0.3 times the other's, both documents were ignored. In addition, when choosing the sentences from the selected documents, the sentences with the length of lower than 8 words did not make it to the final plain texts. With these limitations, the English text contained about 1.4 million sentences and the Persian text one million sentences.

| Documents | | Sentences | |
|---|---|---|---|
| English | Persian | English | Persian |
| 363,183 | 363,183 | 9,933,618 | 1,789,632 |
| 145,479 | 145,479 | 1,391,214 | 1,021,103 |

Table 1: Number of documents and sentences, before (first row) and after (second row) preprocessing

These two plain texts were translated by the initial SMT systems which were trained on Open Parallel Corpus (OPUS) (Lison and Tiedemann, 2016).

## 3.2 Our Method

Our method consists of two main steps: translation and extraction (information retrieval). In translation step we utilize a bidirectional translation approach to extracting parallel sentences from comparable corpora. Two SMT systems are built, one translating from Persian to English and another doing translation from English to Persian, using Moses translation toolkit. Then Lucene IR system is utilized to measure the similarity of sentence pairs. Two similarity scores for a sentence pair are produced by the IR system, one, $Sim_{en-fa}$, for the original Persian sentence and the sentence translated by English-Persian SMT system and another, $Sim_{fa-en}$, for the original English sentence and the one translated by Persian-English SMT system. Based on these two scores, we develop a formula to calculate one similarity score for each sentence pair which is as follows:

$$BiSimilarity = \frac{\alpha}{\alpha + Penalty} \times \frac{\beta \times Sim_{fa-en} + Sim_{en-fa}}{\beta + 1} \quad (1)$$

The coefficient β in the formula represents the logarithm of relative translation quality of one machine against the other. Since the quality of the two SMT systems were different, we decided to assign them different weights in the formula. The quality of the Persian-English system, which is 19.78 by BLEU evaluation metric (Papineni et al., 2002), is almost triple that of English-Persian system (7.80). However, assigning the weight 3 to Persian-English system against 1 to the other one would make English-Persian system almost uninfluential in the process of parallel sentence extraction, hence the weight 1.5 as opposed to 1 which, relatively, are the logarithmic values of the qualities of the two SMT systems. In addition, we penalize the sentences that have more or less number of words than their translated equivalents. Therefore, the Penalty variable, which is the difference in the word number of the two sentences, makes the similarity score smaller when the difference is too large. The average number of words in each sentence in the corpus is α which in our experiments was 22.

### 3.2.1 Translation

Moses toolkit has been widely used for translation in recent years. Therefore, we chose this toolkit for translating our plain texts. The initial systems for translating Wikipedia articles were built on OPUS collection which is a parallel collection of movie subtitles in many languages and is available online for public access. The Persian-English corpus we downloaded consisted of more than 3.7 million sentences in both languages. Three and a half million sentences were used for training, 200 for tuning, and 200,000 for testing. The BLEU scores of the baseline systems were 19.78 and 7.80 for Persian to English and English to Persian, respectively. To build the translation systems, the default settings for Giza++ and SRILM toolkit were used.

### 3.2.2 Information Retrieval

We employed Lucene IR system for extracting parallel sentences. Lucene is a java program which can be used for indexing all the documents in a directory and performing queries on the indexed files. The queries can consist of several words and the results shown by the IR system are the most relevant documents to a given query with a score representing the degree of their relevance. The formula with which Lucene measures the relevance of a document is based on term frequency and inverse document frequency. The documents are ranked with the most relevant as number one and the least relevant at the end. We made use of Lucene to measure the similarity of the translated sentences and the original ones. For each English sentence, 10 Persian candidate sentences were recognized with their similarity scores calculated by the IR system and the same was done for each Persian sentence. Then, using the score for each sentence pair in Formula 1, we chose the candidates that scored the highest. Also we allowed two candidates to be chosen for one sentence when it was possible.

## 4. Experiments

In this section, first some detailed information about the extracted corpus is given, and then the results of several experiments which were conducted on the extracted sentences are presented.

In order to compare the bidirectional method with one-directional method, both methods were implemented which resulted in the extraction of 158339 sentences by one-directional method and 199936 sentences by bidirectional one. The extracted sentences have been sorted by their degree of similarity score calculated by the IR system and Formula 1. The produced scores were divided into 6 intervals to determine the number of sentence pairs that belong to each interval. The result is presented in Figure 2.

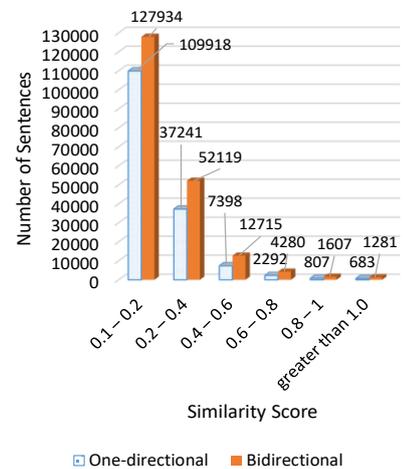

Figure 2: The number of extracted sentences by each method with their similarity scores

| Similarity Score | Extracted Sentences |
|---|---|
| 3.6 | On the eve of his execution he talked of suicide<br>در آستانه اعدام او از قصد خودکشی صحبت کرد |
| 1.5 | Mycobacteria infect many different animals including birds rodents and reptiles<br>میکوباکتری‌ها بسیاری از حیوانات مختلف را آلوده می‌کنند، از جمله پرندگان،جوندگان، و خزندگان |
| 1.0 | The two kinds of variables commonly used in Smalltalk are instance variables and temporary<br>دو نوع متغیر معمول استفاده شده در اسمالتاک متغیرهای نمونه و متغیرهای موقت هستند |
| 0.9 | Hatem was born in Burbank California and grew up in Monterey Park California<br>هتم در بربنک کالیفرنیا بدنیا آمد و در مانتری پارک کالیفرنیا بزرگ شد |
| 0.7 | Today their fragmented and partly degraded range extends from India in the west to China and Southeast Asia<br>امروزه باقی مانده زیستگاه ببرها از هند در غرب تا چین و آسیای جنوب شرقی کشیده شده است |
| 0.5 | It is bordered to the north by the Gulf of Finland to the west by the Baltic Sea to the south by Latvia and to the east by Lake Peipus and Russia<br>این کشور از غرب به دریای بالتیک و از شمال به خلیج فنلاند می‌رسد و مرز شرقی آن با روسیه و مرز جنوبی آن با لتونی مشترک است |
| 0.3 | The difficulty is getting enough data of the right kind to support the particular method<br>دشواری کار ترجمه خودکار، بدست آوردن اطلاعات کافی از نوع صحیح آن برای پشتیبانی روشی خاص است |
| 0.1 | The role of the passive audience therefore has shifted since the birth of New Media and an ever-growing number of participatory users are taking advantage of the interactive opportunities especially on the Internet to create independent content<br>تولید محتوا همچنان در حال رشد است و مصرف محتوا ناشی از رشد موبایل و وسایل دیگر هم در حال رشد و گسترش است |

Table 2: A sample of extracted sentence pairs. In each entry, first sentence is an English sample and the second line is corresponding extracted Persian one

By Looking at Figure 2, it can be observed that there are few sentence pairs with a higher similarity score than 0.4. They make up almost 10 percent of the corpus. Yet, these are the ones that contribute much to the performance of the systems built on this corpus (Figure 3). Although the contribution of the other 90 percent is small, it is still noticeable. In Table 2, a sample of sentences extracted by the proposed method is presented.

We checked the quality of the extracted sentences in two ways: (1) by building an SMT system using only the extracted sentences (Figure 3) and (2) by building a baseline SMT system using 500,000 sentences from OPUS collection and then adding the extracted sentences to the baseline system (Table 3). In both ways, our method performed slightly better than one-directional method.

To tune and test the SMT systems trained on the extracted sentences, we collected 200 sentences (ak-tune-200) for the tuning part and 1000 sentences (ak-test-1k) for testing. Five hundred sentences from the test collection are the ones translated by some colleagues of ours at an English institute. We collected the other 500 and also the 200 sentences of the tuning collection from some websites which offered free parallel sentences. These sentences are all taken from paper abstracts. We proofread all of them one by one to make sure that they have been translated correctly and also made sure that none of them was taken from Wikipedia or movie subtitles. The language model was built by combining Wikipedia documents with OPUS collection.

The first 100,000 sentences extracted by one-directional method have been named 'one-directional-100k' and the ones extracted by our method have been named 'Bidirectional-100k'. Because of the randomness hidden in the tuning phase of an SMT system, every time it is implemented, the result of the translation can be slightly different. In order to obtain more reliable results, we implemented the SMT systems 3 times for each test. Therefore, the BLEU scores shown in the Table 3 are the average of the three scores.

In order to compare the quality of the extracted sentences with the OPUS collection, we built another system using 3.5 million sentences from the OPUS collection. As can be seen in Table 3, its quality was lower than that of the system built by our extracted data although the number of sentences in OPUS3.5M was 35 times higher than that of ours. This can be attributed to the nature of the OPUS collection which is a collection of movie subtitles, making it unable to translate formal sentences with good quality. The number of sentences in the test set can also affect the BLEU score. To show this, we added 4000 more sentences

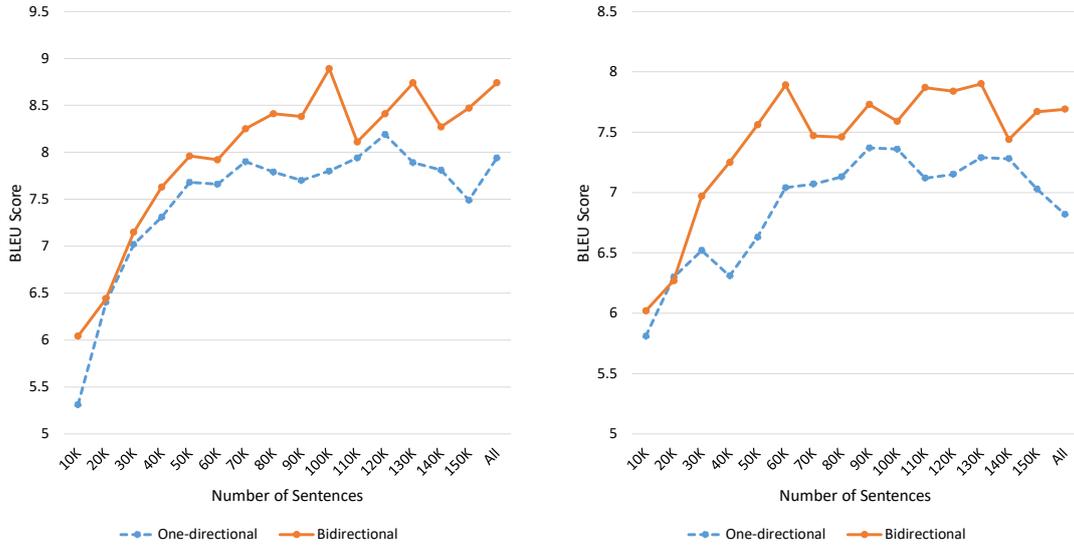

Figure 3: The performances of SMT systems built on various numbers of extracted sentences by both methods. Left: Persian-English system. Right: English-Persian system

to the test set and as is shown for the third test set in Table 3, the BLEU score went up by almost 1.5 points. In addition to Persian-English SMT systems, experiments with English-Persian SMT systems were carried out whose results can be seen in Table 3. As can be seen, the trend for the latter is similar to the former, indicating that the proposed method outperforms the one-directional method in both directions.

| Test Sets | Corpus | BLEU Fa-En | BLEU En-Fa |
|---|---|---|---|
| ak-test-1k | OPUS3.5M | 6.08 | 1.57 |
|  | One-directional-100k | 7.94 | 7.05 |
|  | **Bidirectional-100k** | **8.39** | **7.58** |
| ak-test-1k + OPUS1k | OPUS500k | 6.22 | 3.54 |
|  | OPUS500k + One-directional-100k | 9.92 | 7.86 |
|  | **OPUS500k + Bidirectional-100k** | **10.21** | **8.23** |
| ak-test-1k + OPUS5k | OPUS500k | 9.11 | 5.26 |
|  | OPUS500k + One-directional-100k | 11.45 | 7.02 |
|  | **OPUS500k + Bidirectional-100k** | **11.70** | **7.30** |

Table 3: Results of experiments with Persian-English and English-Persian SMT systems using 500k sentences of OPUS collection and 100k sentences of the extracted corpora by one-directional and bidirectional methods

It is worth noting that if the number of sentences from OPUS collection in the test set increases, the quality of English-Persian system is not guaranteed to improve as is the case with our test sentences combined with 5000 thousand sentences from OPUS. In this case, although the Persian-English system translates better than previous ones, the quality of English-Persian system drops. One way for this anomaly to be explained is by looking at the Persian side of our extracted corpus and that of OPUS corpus. Since the former is used for language modeling and is significantly different from the latter in terms of grammatical structure and the use of words, not to mention the inconsistencies prevalent in the typesetting of the OPUS collection, when more sentences from OPUS are added to the test set, the system's quality deteriorates. This is not the case regarding Persian-English system due to the fact that English side contains less problematic typesetting.

## 5. Conclusion

Parallel corpora are an important part of a statistical machine translation system. However, there is a lack of such data available for everyone. In this paper, a bidirectional method to extract parallel sentences from Wikipedia documents was proposed. The documents were translated from Persian to English and also in the reverse direction in order to find equivalent sentences. Furthermore, a similarity score was proposed to choose the best equivalents. Several different experiments with Persian-English and English-Persian SMT systems were carried out to show the quality of the extracted corpus. It was shown that existing SMT systems performed better when the extracted sentences were added to the systems. It was also demonstrated that the corpus extracted by bidirectional method performs better than the corpus extracted by one-directional approach by approximately 0.5 points in BLEU score. The sentences extracted by both methods have been made available online. As future work, instead of translating the documents by a statistical machine translation system, deep learning models such as word2vec, which are becoming more popular due to their high performance compared to statistical models, can be used for translation.

## 6. Acknowledgements
We would like to thank our colleagues, Zahra Sepehri and Ailar Qaraie, at Iranzamin Language School for providing us with 500 sentences used in our test set.

## 7. Bibliographical References

Abdul-Rauf, S., & Schwenk, H. (2009). On the use of comparable corpora to improve SMT performance. In Proceedings of the 12th Conference of the European Chapter of the Association for Computational Linguistics (pp. 16-23). Association for Computational Linguistics.

Aker, A., Paramita, M., & Gaizauskas, R. (2013). Extracting bilingual terminologies from comparable corpora. In Proceedings of the 51st Annual Meeting of the Association for Computational Linguistics (Volume 1: Long Papers) (Vol. 1, pp. 402-411).

Ansari, E., Sadreddini, M. H., Sheikhalishahi, M., Wallace, R., & Alimardani, F. (2017). Using English as Pivot to Extract Persian-Italian Parallel Sentences from Non-Parallel Corpora. *arXiv preprint arXiv:1701.08339*.

Bakhshaei, S., Khadivi, S., & Safabakhsh, R. (2015). A Generative Model for Extracting Parallel Fragments from Comparable Documents. ACL-IJCNLP 2015, 43.

Do, T. N. D., Besacier, L., & Castelli, E. (2010). A fully unsupervised approach for mining parallel data from comparable corpora. In European COnference on Machine Translation (EAMT) 2010 (p. xx).

Do Thi Ngoc Diep, L. B., & Castelli, E. Improved Vietnamese-French Parallel Corpus Mining Using English Language.

Fung, P., & Cheung, P. (2004). Mining Very-Non-Parallel Corpora: Parallel Sentence and Lexicon Extraction via Bootstrapping and E. In EMNLP (pp. 57-63).

Koehn, P. (2005). Europarl: A parallel corpus for statistical machine translation. In MT summit (Vol. 5, pp. 79-86).

Koehn, P., Hoang, H., Birch, A., Callison-Burch, C., Federico, M., Bertoldi, N., ... & Dyer, C. (2007). Moses: Open source toolkit for statistical machine translation. In Proceedings of the 45th annual meeting of the ACL on interactive poster and demonstration sessions (pp. 177-180). Association for Computational Linguistics.

Li, B., & Gaussier, E. (2010). Improving corpus comparability for bilingual lexicon extraction from comparable corpora. In Proceedings of the 23rd International Conference on Computational Linguistics (pp. 644-652). Association for Computational Linguistics.

Linard, A., Daille, B., & Morin, E. (2015). Attempting to bypass alignment from comparable corpora via pivot language. ACL-IJCNLP 2015, 32.

Lison, P., & Tiedemann, J. (2016). OpenSubtitles2016: Extracting Large Parallel Corpora from Movie and TV Subtitles. In LREC.

Munteanu, D. S., & Marcu, D. (2005). Improving machine translation performance by exploiting non-parallel corpora. Computational Linguistics, 31(4), 477-504.

Munteanu, D. S., & Marcu, D. (2006, July). Extracting parallel sub-sentential fragments from non-parallel corpora. In Proceedings of the 21st International Conference on Computational Linguistics and the 44th annual meeting of the Association for Computational Linguistics (pp. 81-88). Association for Computational Linguistics.

Papineni, K., Roukos, S., Ward, T., & Zhu, W. J. (2002, July). BLEU: a method for automatic evaluation of machine translation. In Proceedings of the 40th annual meeting on association for computational linguistics (pp. 311-318). Association for Computational Linguistics.

Rauf, S. A., & Schwenk, H. (2011). Parallel sentence generation from comparable corpora for improved SMT. Machine translation, 25(4), 341-375.

Resnik, P., & Smith, N. A. (2003). The web as a parallel corpus. Computational Linguistics, 29(3), 349-380.

Seo, H. W., Cheon, M. A., & Kim, J. H. (2015). Extracting Bilingual Lexica from Comparable Corpora Using Self-Organizing Maps. ACL-IJCNLP 2015, 62.

Smith, J. R., Quirk, C., & Toutanova, K. (2010). Extracting parallel sentences from comparable corpora using document level alignment. In Human Language Technologies: The 2010 Annual Conference of the North American Chapter of the Association for Computational Linguistics (pp. 403-411). Association for Computational Linguistics.

Ştefănescu, D., & Ion, R. (2013). Parallel-Wiki: A collection of parallel sentences extracted from Wikipedia. In Proceedings of the 14th International Conference on Intelligent Text Processing and Computational Linguistics (CICLING 2013) (pp. 24-30).